\documentclass{article}
\pdfoutput=1

\usepackage{arxiv}

\usepackage[utf8]{inputenc} % allow utf-8 input
\usepackage[T1]{fontenc}    % use 8-bit T1 fonts
\usepackage{hyperref}       % hyperlinks
\usepackage{url}            % simple URL typesetting
\usepackage{booktabs}       % professional-quality tables
\usepackage{amsfonts}       % blackboard math symbols
\usepackage{nicefrac}       % compact symbols for 1/2, etc.
\usepackage{microtype}      % microtypography
\usepackage{lipsum}		% Can be removed after putting your text content
\usepackage{graphicx}
\usepackage{doi}

\usepackage{amssymb}
\usepackage{amsmath}
\usepackage{amsthm}
\usepackage{lineno}

\usepackage{multirow}
\usepackage{multicol}
\usepackage{makecell}
\usepackage{tabularx}
\usepackage{breqn}

\usepackage{mathtools}
\usepackage{array}
\newcolumntype{C}{>{$}c<{$}}

\usepackage[ruled,vlined,noend,linesnumbered]{algorithm2e}
\usepackage{caption}

\title{AEFE: Automatic Embedded Feature Engineering For Categorical Features}

\author{
Zhenyuan Zhong$^{12}$ \hspace{0.1em}
Jie yang$^{12}$ \hspace{0.1em}
Yacong Ma$^{12}$ \hspace{0.1em} 
\AND
Shoubin Dong$^{12}$\thanks{Corresponding author} \hspace{0.1em}
Jinlong Hu$^{12}$ \\

\\
$^1$ School of Computer Science and Engineering\\
South China University of Technology\\
Guangzhou 510006, China\\
\\
$^2$ Guangdong Key Laboratory of Communication and Computer Network\\
South China University of Technology\\
Guangzhou 510006, China \\
\\

\hspace{0.5em} \{crfcolf, jasonyang8119\}@gmail.com, yacongma@foxmail.com, \{sbdong, jlhu\}@scut.edu.cn \\
}

\renewcommand{\shorttitle}{AEFE: Automatic Embedded Feature Engineering For Categorical Features}

\hypersetup{
pdftitle={AEFE: Automatic Embedded Feature Engineering For Categorical Features},
pdfsubject={IR, ML},
pdfauthor={Zhenyuan Zhong, Jie Yang},
pdfkeywords={utomatic feature engineering, Click-through rate prediction, Combinatorial features, Feature selection},
}

\begin{document}
\maketitle

\begin{abstract}
The challenge of solving data mining problems in e-commerce applications such as recommendation system (RS) and click-through rate (CTR) prediction is how to make inferences by constructing combinatorial features from a large number of categorical features while preserving the interpretability of the method. 

In this paper, we propose \emph{Automatic Embedded Feature Engineering}(\emph{AEFE}), an automatic feature engineering framework for representing categorical features, which consists of various components including custom paradigm feature construction and multiple feature selection. By selecting the potential field pairs intelligently and generating a series of interpretable combinatorial features, our framework can provide a set of unseen generated features for enhancing model performance and then assist data analysts in discovering the feature importance for particular data mining tasks. Furthermore, AEFE is distributed implemented by task-parallelism, data sampling, and searching schema based on Matrix Factorization field combination, to optimize the performance and enhance the efficiency and scalability of the framework. Experiments conducted on some typical e-commerce datasets indicate that our method outperforms the classical machine learning models and state-of-the-art deep learning models.
\end{abstract}


\section{Introduction}
\label{sec:introduction}
In general machine learning problems, \emph{feature engineering} (FE)  is a crucial step that influences the final performance. Studies have shown that feature engineering is often more important and time-consuming than models training \cite{domingos2012few}. Automatic feature engineering frameworks may be used to automate this process to saves a considerable amount of time and discover more novel features. Nevertheless, just like the No Free Lunch theorem of model selection \cite{wolpert1996lackNFL}, there is no universally applicable automatic feature engineering solution. Another focused area is deep learning, which emphasizes end-to-end learning. In this area, feature engineering in traditional machine learning is replaced by representation learning. However, there is no ubiquitous deep learning model for all tasks, which means different feature extraction layers need to be designed for different original features.

Some important tasks of data mining, such as recommendation system and click-through rate prediction, usually contain a large number of categorical fields\footnote{Categorical field and categorical feature, which are from the perspective of data form and machine learning respectively, are similar in most contexts of this paper. } (e.g., gender, occupation), each field contains a variety of values (occupation including students, programmers, etc.). A field containing $d_i$ values is converted to $d_i$ binary features via one-hot encoding. In a dataset with $m$ categorical fields, the total number of features is $N_{f}=\sum ^{m}_{i=1}d_i$($N_{f}\gg m$), but the only $m$ of them is $1$ while the rest is $0$, which causes high-dimensional sparsity.

Therefore, how to effectively represent categorical features is a core task of these applications. Among various features, the \emph{combinatorial feature}, which is a combination of different raw features through a certain paradigm, or the \emph{crossing feature} learned by models, is the most crucial. For example, in the advertisement CTR problem, a specific user prefers to click on the snack advertisement in the afternoon since he or she becomes hungry at tea time. 
Shan \cite{shan2016deepcross} believed that they are often the strongest features of many models, while Cheng \cite{cheng2016wideWANDD} also emphasized that they added plenty of manual crossing features in the wide part. The crossing feature "\texttt{times=afternoon\_AND\_ADCategory=snacks}"  brings a more reasonable inference than the one-hot encoded feature  "\texttt{times=afternoon}" and "\texttt{ADCategory=snacks}" separately. The combinatorial features can bring great precision improvement to prediction; Its high interpretability is helpful to dig deep into the underlying relationship of the data. In order to reduce the time-consuming artificial feature engineering of combinatorial features and improve the prediction accuracy, researchers and practitioners invested a lot of time and effort into this aspect. The research substantially focuses on the improvement of machine learning models and automatic feature engineering.

To learn the crossing features, some machine learning models such as Factorization Machines (FM) \cite{rendle2010factorization} represent categorical features as latent vectors and model the feature crossing by the inner product of latent vectors. Inspired by the successful use of deep learning in computer vision and natural language processing, studies \cite{zhang2016FNN, Guo2017DeepFM, he2017NFM} implicitly map categorical features into embedding vectors, use FM to train the embedding layer, and then use Deep Neural Networks (DNN) to obtain higher-order nonlinear interactions of features. The key point of these studies is to design a novel feature crossing layer, So a technique named Neural architecture Search (NAS) from AutoML was applied to design it automatically \cite{liuautofis2020, kfaefe2020, lbautogroup2020}.  Although how to handle the high-dimensional sparse features is solved by deep learning models, they cannot make full use of all active combinatorial features and have less interpretability.

The automatic feature engineering frameworks can also construct specific features. Some of them are oriented to various data mining tasks. They can perform feature engineering on different types of features (e.g., categorical features, numerical features) while maintaining acceptable interpretability. For example, Data Science Machine(DSM) \cite{kanter2015deepDSM}, which iteratively constructs combinatorial features in relational data tables. Based on DSM,  The follow-up works emphasize a more complex construction paradigm \cite{zhang2018AFEM} and the broader data form  \cite{lam2017oneBM}. However, there are neither in-depth studies of combinations for categorical features nor attempts to embed high-dimensional sparse categorical features by automatic feature engineering techniques.

In this paper, we propose \emph{Automatic Embedded Feature Engineering(AEFE)}, which can perform feature engineering on sparse categorical features automatically. It generates complex but interpretable combinatorial features, as shown in Table \ref{table:feature_scheme}, and requires no manual intervention. AEFE is generally applicable to most Internet data mining tasks that contain a large number of categorical features.

Overall, the contribution of this paper is as follows:
\begin{itemize}
    \item We attempt to apply automatic feature engineering technology on the feature combination problem of categorical features and devise AEFE, a novel solution to represent combinatorial features in machine learning.
    \item Rich feature construction paradigms and operation sets are provided by AEFE to generate various meaningful statistical features, which can significantly enhance model performance. Meanwhile, the multi-module cascaded feature selection unit can select the most useful features efficiently.
    \item Diverse optimization techniques like data sampling, field combination search, and distributed implementation are leveraged to ensure that the framework is scalable and efficient. 
    \item By experiments on several typical datasets, it is empirically demonstrated that AEFE cascaded with GBDT(Gradient Boosting Desicion Tree) achieves better performance than state-of-the-art deep learning models. In addition, AEFE has a stronger capture capability of combination information, as well as better interpretability, than some learning models such as FM, AFM, DeepFM, xDeepFM.
\end{itemize}

\begin{table}[ht]
    \caption{Some examples of complex constructed features in different scenarios.}
    \label{table:feature_scheme}
    \centering
    \small
    \begin{tabular}{@{}ll@{}}
    \toprule
    Scenario                         & Combinatorial feature                                                    \\ \midrule
    Item recommendation         & Number of times user $u_j$ collected item in the last 10 days                        \\
    O2O coupon usage prediction      & Number of times user $u_j$ receives coupons of merchant $m_k$                             \\
    Purchase prediction              & \makecell[tl]{Difference between the price of the commodity and \\the average price of commodities in the same\\ category $c_i$ that user $u_j$ has purchased} \\
    Display advertisement prediction & Proportion of video $v_k$ in user $u_j$'s records                              \\ \bottomrule
    \end{tabular}
\end{table}

The remainder of the paper is organized as follows. In Section \ref{sec:relatedwork}, relevant previous works about machine learning models and automatic feature engineering are introduced. In Section \ref{sec:framework}, we present our AEFE in detail and illustrate it from different perspectives. In Section \ref{sec:OptimizationImplementation}, some optimization techniques of AEFE are described. We conduct several experiments in Section \ref{sec:Experiment} to show the performance of our methods, and finally, a brief conclusion and prospects are presented in Section \ref{sec:conclusion}.

\section{Related Work}
\label{sec:relatedwork}

\subsection{Feature Crossing by Models architectures}
\label{subsec:featurecrossinginmachinelearning}

The early feature crossing technique applied to high-dimensional sparse features is derived from the optimization of the binomial kernel method  \cite{kudo2003fast}. Subsequently, POLY2 \cite{chang2010poly2}  is proposed to use the binomial transformation to achieve feature crossing. However, due to the high feature dimension, such models encounter the problem of excessive parameter space and over-fitting.

To alleviate the excessively large parameter space of the crossing feature, FM \cite{rendle2010factorization} attaches the weight of the crossing feature to the inner product between the latent vectors of corresponding features. Owing to its pleasing effect, improving FM becomes an important research direction. Higher-order Factorization Machines (HOFM) \cite{blondel2016HOFM} extends the matrix factorization of FM to tensor factorization to achieve higher-order feature crossing. Xiao \cite{xiao2017AFM} apply the attention mechanism to obtain the weight of second-order terms and propose Attentional Factorization Machines(AFM). Field-aware Factorization Machines(FFM) \cite{juan2016fieldFFM} uses field information additionally to improve the prediction effect. Each feature uses different latent vectors to calculate the interaction with corresponding fields. Pan \cite{pan2018FwFM}  proposes Field-weighed Factorization Machine(FwFM) to model the weights of field pairs, reducing the number of parameters compared with FFM.

Furthermore, FM can be regarded as a network model with only an embedding layer and a hidden crossing layer \cite{zhou2018DIN}. From this perspective, FM may be used as an approach to train the embedding layer, which not only maps high-dimensional sparse features to dense vectors but also captures second-order feature crossing.  Zhang et al. \cite{zhang2016FNN} proposed Factorization-Machine Supported Neural Networks (FNN), which pre-train the embedding vectors of features with FM and feed them into DNN next.  Similar to FNN, Neural Factorization Machines (NFM) uses the Bi-interaction layer to subdivide the inner product of FM, generalizes FM with a deep learning structure \cite{he2017NFM}. Deepfm \cite{Guo2017DeepFM} replaced the wide part of the Wide\&Deep model with FM. The commonness of these models is that after embedding the categorical features, the second-order crossing information and the high-order nonlinear crossing information are captured by FM and DNN, respectively. These models may be referred to as "FM-embedding deep learning models".

Training the embedding vector containing feature crossing information does not necessarily require the use of FM. For example, Product-based Neural Network (PNN) replaces the FM layer with a product layer, which captures different crossing patterns by inner product or outer product \cite{qu2016PNN}. Besides, Deep\&Cross Network (DCN) \cite{wang2017DCN} and eXtreme Deep Factorization Machines (xDeepFM) \cite{lian2018xdeepfm}, utilize Cross Net and Compress Interaction Network (CIN) respectively to learn specific orders of explicit feature crossing, while the former interact features at the bit-wise level and the latter at the vector-wise level. XDeepInt \cite{yanxdeepint2020} then ensemble a Polynomial Interaction Network (PIN) to learn both vector-wise and bit-wise level feature interactions. In addition, it is acceptable to use CNN \cite{liu2015CCPM,chan2018MSS} or directly use DNN \cite{cheng2016wideWANDD,zhou2018DIN} to model feature interaction. What's more, extensive research proposed various models to learn feature crossing, e.g., Deep Crossing \cite{shan2016deepcross}, Deep and Shallow Layers (DSL) \cite{huang2017DSL}, and Fi-GNN \cite{lifignn2019} which applied Graph Neural Network, just to name a few.

The models mentioned above can learn different order feature crossing, especially the state-of-the-art deep learning models that can capture non-linear crossing. However, the large amount of parameters makes them only suitable for applications with a large amount of data. Moreover, they have problems of low prediction stability and difficult tuning, compared with the pattern of feature engineering plus traditional machine learning. To handle it, recently, NAS was used to find appropriate layers for different feature interactions. It focuses on leveraging mining of the high-order feature interactions and the complexity of model architectures. Inspired by DARTs \cite{liu2018darts}, a novel algorithm for differentiable NAS, AutoFIS \cite{liuautofis2020} introduce a gate for each feature interaction on FM layer in training procedure to find useful combinational features in inference step. AutoFeature \cite{kfaefe2020} then explores feature interactions on each order of feature fields based on their pre-defined operations including Pointwise Addition, Hadamard Product, Concatenation, Generalized Product, and Null. On the contrary, AutoGroup \cite{lbautogroup2020} defines a fixed feature interaction function extending from FM but explores which feature is useful in which order interaction.

The application of NAS hasn't changed the functions of basic feature interaction. A benchmark conducted by Zhu \cite{zhu2020fuxictr} indicates that designing a novel architecture meets the bottleneck in pursuing better performance. Besides, either the implicit crossing feature learned by DNN or the explicit crossing feature learned by PNN, CIN, GNN, PIN, etc., makes the model's interpretability weak, which leads to the fact that in some business scenarios, the prediction results cannot give valuable feedback to the data collection and preprocessing stages. Taking the online recommendation system as an example, the deep learning model can give accurate recommendation items, but it is difficult to know which context information or what combinatorial features are working to select this item. Guo \cite{guo2018visualizing} conducts experiments to explore the understanding of DNN on CTR prediction problems through visualization, yet its work mainly analyzes the internal mechanism of DNN and gives no guiding analysis from the business level.

\subsection{Automatic Feature Engineering}
\label{subsec:automaticfeatureengineering}

The study of automatic feature engineering and deep learning models is quite different, but they are similar in the goals of reducing time-consuming manual feature engineering and improving prediction accuracy just like NAS.

Some of the early automatic feature engineering studies mostly focus on finding effective transformation functions, which are applied to the appropriate features, thus improving the prediction effect \cite{sondhi2009survey}. FICUS \cite{markovitch2002feature} uses beam search to find features in the construction space, and takes advantage of Information Gain (IG) as a proxy function to judge whether the feature is valid. FCtree \cite{fan2010FCTree} also uses IG to filter the generated features. It trains the decision tree model and selects the structural features that can bring the gain at the same time. Unlike the way we attach weight to field combination, FCTree assigns weights to the transformation function and constantly updates them. FEADIS randomly selects features to construct and uses cross-validation to verify the efficacy of features, which can be regarded as a greedy strategy \cite{dor2012strengthening}. What the above work has in common is that 1) feature construction only uses data of a single sample, 2) the construction paradigm is simple, and 3) they only apply to small datasets.

Some studies attempt to use the learning model to improve automation and enhancement in the process of automatic feature engineering.  ExploreKit \cite{katz2016explorekit} uses a ranking model to select the optimal constructor. Nargesian \cite{nargesian2017LFE} defines whether to choose a particular transformation as a binary classification task, so a Multilayer Perceptron (MLP) was trained on the meta-features of plenty datasets for prediction. It is worth noting that the above two studies adopt meta-learning in their framework. Besides, AutoLearn \cite{kaul2017autolearn} automatically constructs two kinds of features with a regression model, which makes the machine learning task improve obviously. Our work integrates a gradient descent process of Matrix Factorization (MF) into the field combination search, which improves efficiency.

Cognito \cite{khurana2016cognito} expresses the automatic feature problem as a tree-structured task: Each node is a dataset that joins the construction features, and each edge is a transformation function.  AutoCross \cite{yuanfeiautocross2019} shared the same viewpoint but didn't focus on feature engineering. Khurana \cite{khurana2018FE-RL} further describes the task as a Directed Acyclic Graph (DAG) and uses reinforcement learning to trade off factors like effectiveness and efficiency. These frameworks can iteratively use transformation functions to make the constructed features more complex. Features complicated by iterative transformation, while improving the effectiveness, are more difficult to understand. How to explain the meanings of a feature after multiple conversions?

In order to construct more valuable features, Data Science Machine (DSM) \cite{kanter2015deepDSM} makes use of more samples information by using aggregation operations (Groupby) in the relational table, which makes DSM more practical in data mining tasks. Lam's OneBM  \cite{lam2017oneBM} improved DSM and added support for unstructured data. AFEM's improvements focused on complex feature paradigms, which defined multiple families of features such as the family of social graph-based features \cite{zhang2018AFEM}. These work for realistic data mining scenarios are more practical than some previous works.

The above works consider constructing different forms of features with as many data types as possible to provide universality. However, the actual data mining tasks vary widely, so the unified solution has a limited effect, and it is challenging to introduce prior knowledge guidance. Furthermore, for most of the automatic feature engineering research, the construction process only uses the feature information, yet the label information is only used for verifying the validity of features, which makes the information captured by these frameworks different from that of the deep learning model (since deep learning model directly fits the relationship between features and labels).

The interpretability of features or models, the complexity of finding high-order feature interactions, and the capability of handling high-dimensional sparse features are all points that need to be considered. To the best of our knowledge, most existing related works can only satisfy partial demands mentioned above, for example, deep learning models \cite{Guo2017DeepFM, he2017NFM, lian2018xdeepfm} loss the interpretability of the features, and some automatic feature engineering framework \cite{kanter2015deepDSM, khurana2016cognito, lam2017oneBM} fail to generate combinatorial features from categorical features. The pros and cons of different methods are summarized in Table \ref{table:comparison_method}.

\begin{table}[ht]
    \caption{Comparison of some related works.}
    \label{table:comparison_method}
    \centering
    \footnotesize
    \begin{tabularx}{\linewidth}{X p{2.8cm} p{2.8cm} p{2.3cm}}
    \toprule
    \textbf{Methods} & \textbf{Complexity of feature interaction} & \textbf{Combination of categorical features} & \textbf{Interpretability} \\
    \midrule
    \textbf{Feature Crossing by Model Architectures}    &    &    &  \\
    Traditional machine learning models, e.g.,\newline{}FM \cite{rendle2010factorization} \newline{}FFM \cite{juan2016fieldFFM} & low   & $\checkmark$   & weak \\
    Deep Learning models, e.g.,\newline{}NFM \cite{he2017NFM} \newline{}xDeepFM \cite{lian2018xdeepfm} & high  & $\checkmark$   & weak \\
    NAS-based models, e.g.,\newline{}AutoFeature \cite{kfaefe2020} \newline{}AutoGroup \cite{lbautogroup2020} & high  & $\times$     & weak \\
    \textbf{Automatic Feature Engineering} &       &       &  \\
    DSM \cite{kanter2015deepDSM}  & high  & $\times$     & strong \\
    oneBM \cite{lam2017oneBM} & high  & $\times$     & strong \\
    \textbf{AEFE(our work)} & \textbf{high} & \textbf{$\checkmark$ } & \textbf{strong} \\
    \bottomrule
    \end{tabularx}
\end{table}

\section{Framework}
\label{sec:framework}
The proposed AEFE focuses on categorical data that is not highlighted in other studies. For specific tasks, users can specify a series of custom options on background knowledge to obtain generated features. In addition, partial features constructed by AEFE revealed the relationship between categorical features and prediction target, so that it can also capture information similar to that learned by models. In other words, our framework explores the answers to the following problem:
\begin{enumerate}
    \item How to generate features automatically on a large dataset with high-dimensional sparse features?
    \item How to design universal paradigms to exploit the potential of combinatorial features fully?
    \item  How to ensure the automatic feature engineering framework to work efficiently?
    \item How to maintain the interpretability of the method to benefit data analysis?
\end{enumerate}

\subsection{Problem Definition}
\label{subsec:Problemdefinition}

First, we generalize the machine learning problem to be solved: Given a dataset containing specific forms of data, our goal is to automatically transform the raw features into effective combinatorial features to enhance model performance. More formally, we have:
\begin{itemize}
    \item Categorical field set $\mathcal{F} = \{F_1, F_2, \cdots, F_m\}$, which is the dominant features in the dataset. $m$ is the number of categorical fields.
    \item Indicator set $\mathcal{I}$, including:
    \begin{itemize}
        \item Predicted target $y$, a.k.a \emph{label} of a sample ;
        \item Time value $TS$ (optional);
        \item Other continuous fields$\{I_1, I_2, \cdots, I_{n_I}\}$ (optional). $n_I$ is the number of other continuous fields.
    \end{itemize}
    \item A dataset $\mathcal{D}=(\mathbf{X},\mathbf{Y})$, where $\mathbf{X}=\{\mathbf{x}_i\}_n, \mathbf{y}=\{y_i\}_n$. Each sample contains the above categorical field and optional continuous fields $\mathbf{x}^i=(x^i_{F_1},\cdots,x^i_{F_m},x^i_{TS},x^i_{I_1},\cdots,x^i_{I_l})$, and a label $y_i$ ;
    \item A custom framework parameter collection $\mathcal{H}$, including:
    \begin{itemize}
        \item Time windows $\mathcal{W}=\{W_1,W_2,\cdots,W_{n_w}\}$;
        \item Operator set $\mathcal{O}=\{O_1,O_2,\cdots,O_{n_o}\}$;
        \item Construction paradigms $\mathcal{P}=\{P_1,P_2,\cdots,P_{n_p}\}$;
    \end{itemize}
    $n_w,n_o,n_p$ are the number of corresponding sets respectively.
\end{itemize}
Then, the role of our proposed framework is to construct a batch of features $\mathcal{G}=\{g_1,\cdots,g_d\}$ using the above information, and convert the raw dataset into a generated dataset:
\begin{equation*}
    AEFE(\mathcal{D,H}) =\mathcal{D'}     
\end{equation*}
The features of the generated samples are $\mathbf{x'}^i=({x'}^i_{g_1},{x'}^i_{g_2},\cdots,{x'}^i_{g_d})$. The meaning of these symbols and the workflow of the framework will be described in the following sections.

\begin{figure}[t!p]
    \centering
    \includegraphics[width=1.0\linewidth]{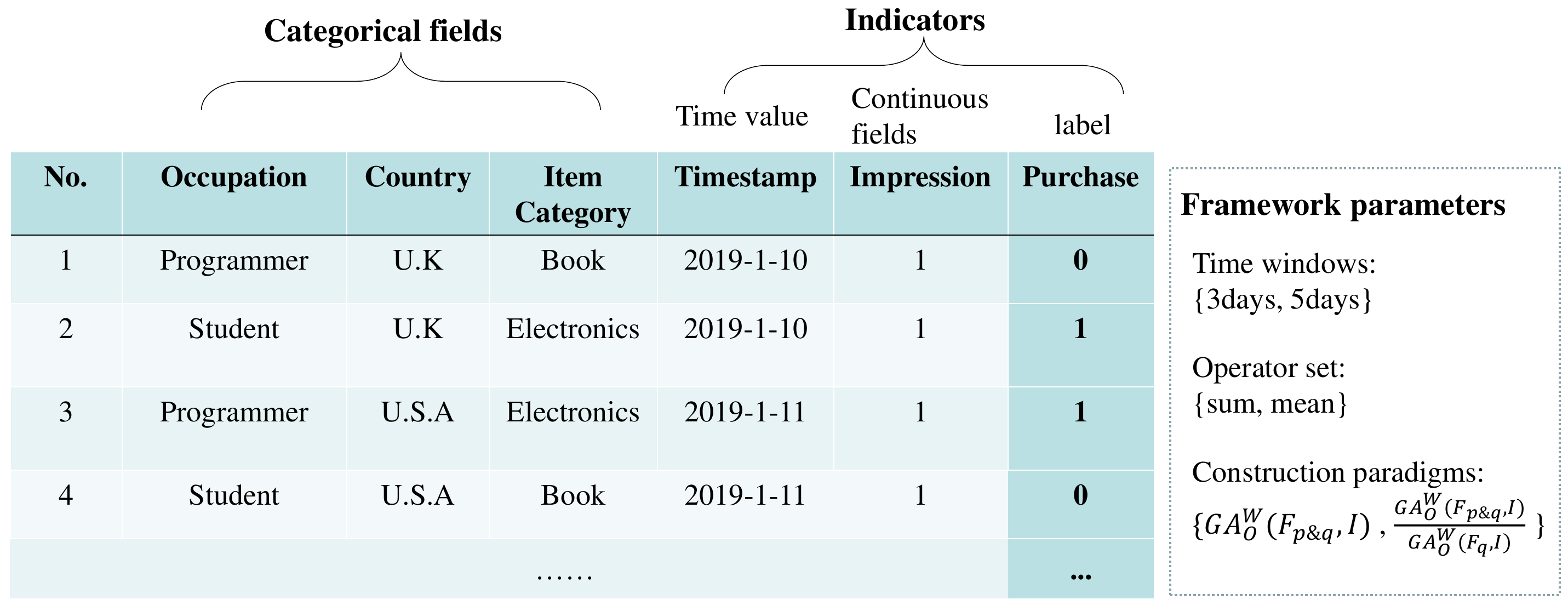}
    \caption{The structure of an E-commerce website purchase prediction dataset and the parameters setting.}
    \label{fig:purchase_prediction}
\end{figure}

We take purchase prediction on an e-commerce website as an example to illustrate the data form mentioned above. As shown in Figure \ref{fig:purchase_prediction}, it is a task to predict whether a user will purchase a specific item by exploiting user information and item information. This dataset includes three categorical fields---\emph{Occupation, Country}, \emph{Item category} and three indicators---\emph{Timestamp, Impression}, and \emph{Purchase}. \emph{Timestamp} represents the time a user viewed the item. \emph{Impression} indicates whether the user viewed the item, and \emph{Purchase} is the predicted target, take 0 or 1. The meaning of the framework parameters is explained in the following sections.

\subsection{AEFE Framework Overview}
\label{subsec:AEFEframeworkoverview}

\begin{figure}[tp]
    \centering
    \includegraphics[width=\linewidth]{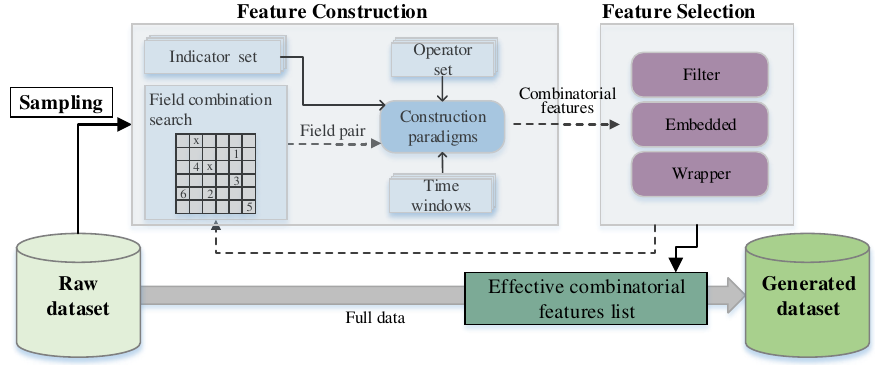}
    \caption{AEFE Overview. It shows how the framework transforms the raw dataset to one with generated features.}
    \label{fig:aefe_overview}
\end{figure}

\begin{table}[ht]
    \caption{Brief description of main modules in \emph{AEFE}}
    \label{table:module_description}
    \centering
    \begin{tabularx}{\linewidth}{p{3.5cm} X}
        \toprule
        Module                 & Description \\ \midrule
        \makecell[tl]{Field combination\\ search} & A search algorithm that uses feedback from the previous iteration to select a field pair based on specific criteria. \\
        Field pair             & Two categorical fields chosen in field combination search, for recalling samples with the same value of these fields.  \\
        Indicator set          & A set of indicators for statistical calculation, e.g. \emph{Impression} and \emph{Timestamp}.  \\
        Operator set           & A set of aggregation functions, e.g. \emph{sum} and \emph{mean}.  \\
        Time windows           & Specific time periods for statistics, e.g. 3 days.  \\
        \makecell[tl]{Construction \\paradigms} & Expressions to determine how to combine various elements. \\
        \makecell[tl]{Combinatorial \\features} & Interpretable continuous features generated by combining various elements of the feature construction phase. Examples are in Table \ref{table:paradigmandfeature}.\\
        Filter                 & A module pre-filting undiscriminate features.  \\
        Embedded               & A module that filters out unimportant features.  \\
        Wrapper                & A module that filters out redundant features.\\ \bottomrule
    \end{tabularx}
\end{table}

An overview of AEFE is shown in Figure \ref{fig:aefe_overview}, and the key modules are explained in Table \ref{table:module_description}. The workflow is as follows: Firstly, a certain proportion of data is \emph{sampled} from the raw dataset, which is used to construct features and verify effectiveness. Secondly,  at the \emph{feature construction} phase, field combination search module generates a categorical field pair based on the feedback information from the previous iteration of feature selection, then various indicators and parameters are coupled to generate \emph{combinatorial features}. Thirdly, these combinatorial features are passed through a multi-module cascaded \emph{feature selection} phase, which filters ineffective features, and the remaining features are added to the \emph{effective combinatorial features list}. Completing the feature construction and feature selection of a field pair is considered to be one iteration. Subsequently, continue to select another field pair and repeat the above process (as indicated by the closed-loop made up by the dotted arrow in the figure) until the field combination search module has no longer generated new field pair. Finally, a complete effective combinatorial features list is obtained, which is used as a template for the full raw data to generate a new dataset.

\subsection{Feature Construction}
\label{subsec:featureconstruction}

\subsubsection{Construction Procedure}
\label{sssec:constructionprocedure}

Feature structure, in essence, is a process of combining categorical fields, indicator sets, operation sets, and time windows to generate combinatorial features specified by the construction paradigm. It can be summarized as follows: 1) Groupby, 2) Aggregating, and 3) Paradigm combination. For sample $\mathbf{x}^k$, the specific structure of feature construction is as follows:
\begin{enumerate}
    \item \emph{Groupby}: After obtaining a field pair $F_p, F_q \in \mathcal{F}$, a set of samples that are equivalent to $\mathbf{x}^k$ on these fields is recalled:
    \begin{equation*}
        S(\mathbf{x}^k \mid F_{p\&q})=\{\mathbf{x}^i \mid x^i_{F_p}=x^k_{F_p} \& x^i_{F_q}=x^k_{F_q}, \quad i=1,\cdots,n,i\neq k\}    
    \end{equation*}
    Or after setting a time window $W_r\in \mathcal{W}$, the sample set is represented as:
    \begin{align*}
        S(\mathbf{x}^k \mid F_{p\&q},W_r)=\{\mathbf{x}^i \mid x^i_{F_p}=x^k_{F_p} \& x^i_{F_q}=x^k_{F_q} \& x^i_{TS}\in (x^k_{TS-W_r},x^k_{TS}), \\
        \quad i=1,\cdots,n,i\neq k\}    
    \end{align*}
    \item \emph{Aggregating}:
    Perform function $O_t$ on indicator $I_s$ of samples in set $S$ to get a scalar $e$:
    \begin{equation*}
        e = O_t(\{x^i_{I_s} \mid \mathbf{x}^i \in S(\mathbf{x}^k)\})
    \end{equation*}
    $O_t$ is an aggregation operator such as \emph{mean} and \emph{sum}. For simplicity, we mark the calculation process of \emph{groupby-then-aggregating} as follows:
    \begin{equation*}
        e = GA^{W_r}_{O_t}(F_{p\&q},I_s)
    \end{equation*}
    \item \emph{Paradigm combination}: Finally, the above calculation result is combined with specified paradigm $P \in \mathcal{P}$ to obtain the construction feature $f_j$, for example:
    \begin{equation}
        P(F,I,W,O|p,q,r,s,t)=\frac{GA^{W_r}_{O_t}(F_{p\&q},I_s)}{GA^{W_r}_{O_t}(F_{p},I_s)} \label{eq:paradigm_example}
    \end{equation}
\end{enumerate}
In the purchase prediction task of Figure \ref{fig:purchase_prediction}, we set $F_p\&F_q$ as \emph{Occupation} \& \emph{Item category}, and use $\mathbf{x}^1$ as an example to clarify the meaning of these steps. In step 1, \emph{Groupby}, all the samples of programmers viewing book items in the last three days---$S(\mathbf{x}^1 \mid Occ.\&Cat.,3ds)$---would be gathered. Then in step 2 \emph{Aggregating}, with function $sum$, the samples sets is aggregated into $GA^{3ds}_{sum}(Occ.\&Cat., Pur.)$ and $GA^{3ds}_{sum}(Cat., Pur.)$, which represent the total amount of books purchased by programmers and the total sales of books in the past three days respectively. By combining these two aggregation values through the paradigm \eqref{eq:paradigm_example}, a combinatorial feature-\emph{proportion of book sales that programmers purchase in the last three days} is obtained in step 3 \emph{Paradigm combination}. More combinatorial features can be generated by using the other framework parameters in Figure \ref{fig:purchase_prediction}. Some examples of features of different paradigms are shown in Table \ref{table:paradigmandfeature}.

\begin{table}[ht]
    \caption{Construction paradigms and corresponding features of the E-commerce purchase scenario.}
    \label{table:paradigmandfeature}
    \centering
    \small
    \begin{tabularx}{\linewidth}{p{3.2cm} p{4.5cm} X}
    \toprule
    Paradigm           & Combinatorial feature                              & Explanation                                                                                                                                 \\ \midrule
    1.$GA^{W_r}_{O_t}(F_{p},I_s)$       & $GA^{d}_{sum}(Occ.,Imp.)$               &  Times user whose occupation is  $F_p^i$ view items in the last $d$ days                         \\
    2.$GA^{W_r}_{O_t}(F_{p\&q},I_s)$      & $GA^{d}_{sum}(Occ.\&Cat.,Pur.)$ & Times user whose occupation is  $F_p^i$ purchase items with category $F_q^i$ in the last $d$ days \\
    3.$\dfrac{GA^{W_r}_{O_t}(F_{p\&q},I_s)}{GA^{W_r}_{O_t}(F_{p},I_s)}$   & $\dfrac{GA^{d}_{sum}(Occ.\&Cat.,Pur.)}{GA^{d}_{sum}(Cat.,Pur.)}$  & Proportion of occupation $F_p^i$'s purchase in sales of category $F_q^i$ in the last $d$ days     \\
    \makecell[tl]{4.$GA^{W_r}_{O_t}(F_{p\&q},I_s)$ \\ $- x^k_I$}  & $GA_{avg}(Cat.,TS) -  x^k_{TS}$        & The difference between the average time of item with category $F_q^i$ sales  and the current time   \\ \bottomrule
    \end{tabularx}
    \begin{center}
        Annotate: \textit{Occ., Cat., Imp., Pur.} and \textit{TS} are short for Occupation, Item Category, Impression, Purchase, and Timestamp.
    \end{center}
\end{table}

\subsubsection{Framework Parameters}
\label{sssec_Frameworkparameters}

Framework parameters include time window, operator set, and construction paradigm. Time windows need to be selected according to the specific situation. Furthermore, the operator set contains the common aggregation function $[sum, mean, std, max, min]$. Lastly, construction paradigms also can be user-defined. However, in this paper, by analyzing the manual feature engineering in various application scenarios, we summarize some general feature construction paradigms, which can be divided into four categories, and they are shown in the column "paradigm" of Table \ref{table:paradigmandfeature} in order. They are defined as:
\begin{enumerate}
    \item \emph{Single-field statistical feature} is a feature obtained by aggregating the historical samples with the same value for a certain field, which can reflect the relationship between a category field and an indicator.
    \item \emph{Multi-field combination statistical feature} is constructed in a similar form to the single-field statistical feature, except that the samples to be computed are determined by two fields. This type of feature can effectively mine effective categorical field combination.
    \item \emph{Nondimensionalization feature}is a multi-field feature divided by a single-field feature, which represents the proportion of an indicator aggregate value in another.
    \item \emph{Distance measurement feature} is the relationship between the single(multi)-field feature and the indicator value of the current sample.
\end{enumerate}

\subsection{Feature Selection}
\label{subsec:featureselection}

\begin{algorithm}[t]

    \SetAlgoLined
    \SetKwInOut{Input}{Input}\SetKwInOut{Output}{Output}
    \SetKwData{lm}{lm}
    % \SetKwData{feature}{feature}
    \SetKwFunction{Std}{Std}\SetKwFunction{Delete}{Delete}\SetKwFunction{SplitData}{SplitData}\SetKwFunction{FeatureImportance}{FeatureImportance}

    % \Input{$\mathcal{D}$, $\mathcal{G}_{FG}$, $pred_{init}$, $rate_{valid}$, $score_{init}$, $t_{filter}$, $t_{embedded}$, $t_{wrapper}$}
    \Input{$\mathcal{D}$, $\mathcal{G}_{FG}$}
    \Output{$\mathcal{G}_{valid}$}

    \For{$g \in \mathcal{G}_{FG}$}{
        \If(\tcp*[f]{Filter module}\label{filter}){\Std{$\mathcal{D}_{g}$}$<t_{filter}$}{
            \Delete $\mathcal{D}_{g}$; $\mathcal{G}_{FG} \leftarrow \mathcal{G}_{FG}-\{g\}$\;
            }
    }
    $data^{tr},data^{va} \leftarrow \SplitData(\mathcal{D},rate_{valid})$\;
    \lm = model(); \lm.fit($data^{tr}$)\;
    \For{$g \in \mathcal{G}_{FG}$}{
        \If(\tcp*[f]{Embedded module}\label{embedded}){$\FeatureImportance(\lm,g)<t_{embedded}$}{
            % \Delete $\mathcal{D}_{f}$\;
            \Delete $data^{tr}_{g},data^{va}_{g}$;  $\mathcal{G}_{FG} \leftarrow \mathcal{G}_{FG}-\{g\}$\;
            }
        }
    Sort $\mathcal{G}_{FG}$ in descending order of feature importance \;
    $\mathcal{S}\leftarrow \emptyset$ \;
    \For{$g \in \mathcal{G}_{FG}$}{
        \lm = model(); \lm.fit($data^{tr}_{\mathcal{S} \cup \{g\}},pred_{init}$)\; 
        pred = \lm.predict(($data^{va}_{\mathcal{S} \cup \{g\}},pred_{init}$)\;
        Calculate $score_{current}$\;
        \If(\tcp*[f]{Wrapper module}\label{wrapper}){$score_{current}-score_{init}>t_{wrapper}$}{
            $score_{init} \leftarrow score_{current}$; Update $pred_{init}$\;
            $\mathcal{S} \leftarrow \mathcal{S} \cup \{g\}$
            }
        \Else{
            $\mathcal{G}_{FG} \leftarrow \mathcal{G}_{FG}-\{g\}$\;
            }
        }
    $\mathcal{G}_{valid} \leftarrow \mathcal{G}_{FG}$\;
    \Return $\mathcal{G}_{valid}$;
    \caption{Feature Selection Algorithm(FSA)\label{alg:FSA}}
\end{algorithm}

A large number of combinatorial features generated in the feature construction phase will bring huge space overhead to the subsequent model training, and the expressiveness of these features may be redundant. Therefore, feature selection of the combinatorial features is needed. The feature selection algorithm consists of three modules: \emph{Filter, Embedded} and \emph{Wrapper}.

\emph{Feature Selection Algorithm} is shown in Algorithm \ref{alg:FSA} and the detailed description of the three modules is as follows:
\begin{enumerate}
    \item \textbf{\textit{Filter}}: It calculates the variance of each feature and drops the features whose variance is less than $t_{filter}$ because features with small variances tend to be less distinguishable.
    \item \textbf{\textit{Embedded}}: The feature is put into a machine learning model capable of calculating feature weights(e.g., GBDT and RandomForest), and that with weight less than $t_{embedded}$ are dropped.  In this way, taking into account the relative importance of all features, features that contribute less to the prediction can be removed. These features with smaller weights may even contain noise, affecting prediction accuracy.
    \item \textbf{\textit{Wrapper}}: Finally, the wrapper module sorts the features in descending order of feature importance. Using the cascaded method, it adds each feature to train the model one by one,  to find out whether the prediction effect can be improved after a feature is added.   The feature is retained if it can bring improvement larger than $t_{wrapper}$; otherwise, it is dropped since it means that the current feature has redundancy with the selected feature set. The procedure terminates when all features have been tried.
\end{enumerate}

An ordered combination of multiple feature selection strategies can complement each other. \emph{Filter} has the highest efficiency and can quickly filter out the features with lower discrimination because it does not need to train the model. However, it is not enough to use the variance to judge whether the feature is potent. \emph{Embedded} takes into account specific prediction tasks and with acceptable speed. With the view of diminishing the features redundancy and substantially scale down the final number of features, the most rigorous but time-consuming method --- \emph{Wrapper}, is used in the last step. Since some features have been dropped in the first two steps, the overall time of this step is acceptable.

\subsection{Two Perspectives on AEFE}
\label{subsec:twoperspectives}
In the previous section, the entire framework has been described in detail. Here, we illustrate AEFE from two perspectives. One perspective is as a work in the AutoML domain, which is the characteristic of AEFE. Another perspective is the difference between AEFE and machine learning methods as a feature extractor.

\subsubsection{AutoML Perspective}
\label{sssec:automlpers}

In the study of AutoML, AEFE is a work for automating the feature engineering phase, which is characterized by constructing continuous combinatorial features for multi-field categorical features. As \cite{quanming2018AutoML} mentioned, most of the AutoML problems can be defined as iteratively using \emph{Optimizer} search \emph{configuration} on a specific \emph{Search space} and then evaluating the feedback with \emph{Evaluator} to complete the optimization task. Here, we interpret the meaning of AEFE in AutoML form:

\begin{itemize}
    \item Search space: search space is determined by the combination of all possible generated features if the number of generated features is $Q$:
    \begin{align}
        \centering
        Q&=|\mathcal{F}|^2/2\times |\mathcal{I}|\times |\mathcal{W}|\times |\mathcal{O}|\times |\mathcal{P}| \label{eq:feature_num}
    \end{align}
    Then the size of the entire search space is $N_{space}=\sum^Q_{i=1} Comb(i,Q)$, where $Comb(i,Q)$ means $Q$ choose $i$.
    \item Optimizer: A greedy search optimization strategy is used in generating and selecting feature sets phase, which is embodied in two aspects. One is to select the current optimal field pair in field combination search, and the other is the criterion of feature selection---whether a feature can bring improvement in the current stage.
    \item Evaluator: On the one hand, the evaluation techniques used in this work include data sampling and a direct training model to evaluation. On the other hand, after feature selection, the feedback information is transmitted back to the optimizer by means of proxy evaluation.
\end{itemize}
In general, due to the large search space, we use tightly coupled tools and techniques in the process of Optimize-Evaluate to improve the efficiency of the proposed framework. Among them, the field combination search module plays an important role in this process. See Section \ref{subsec:mf-basedFCS} for internal optimization details.

\subsubsection{Feature Extractor Perspective}
\label{sssec:featureextractorpers}

\begin{figure}[tp]
    \centering
    \includegraphics[width=\linewidth]{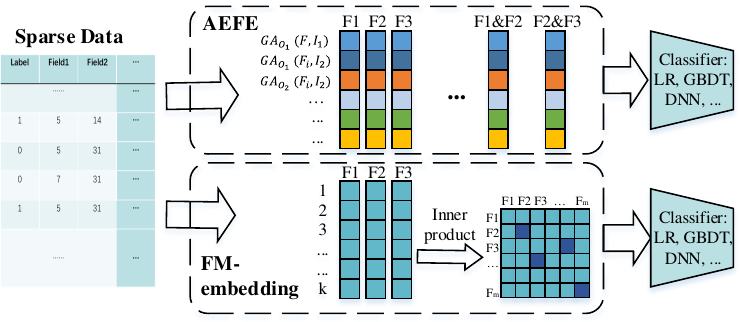}
    \caption{AEFE vs. FM-embedding. They use different techniques to capture feature combinations.}
    \label{fig:aefevsfm}
\end{figure}

As a feature extractor, AEFE converts high-dimensional sparse features into continuous features with practical meanings. This characteristic that is similar in form to representation learning is the essence of AEFE's \emph{"Embedded"}.

When the indicator and operator used by AEFE are limited to \emph{label} and \emph{sum/mean}, it can be considered as a method to extracts the relationship between the combination of categorical features and \emph{label}, from the perspective of the data information utilized, which is consistent with FM. From the aspect of the output form, AEFE and FM both generate a series of continuous variables. The difference is that AEFE performs aggregation operations on the training samples, and FM uses these samples to update parameters through optimization such as stochastic gradient descent(SGD). The similarities and differences between AEFE and FM are shown in Figure  \ref{fig:aefevsfm}. In the classification task, the expression of FM is:
\begin{equation}
    FM(x) = \sigma(\sum\limits_{i=1}^n w_i x_i + \sum\limits_{i=1}^{n-1}\sum\limits_{j=i+1}^{n}<v_i,v_j>x_ix_j)
\end{equation}
$\sigma(x)$ means Logistic Regression(LR). Therefore, FM can be decomposed into obtaining the latent vectors by FM-embedding, get combinatorial features through inner product, and then LR is used as the classifier. From this perspective, AEFE, like FM-embedding, plays the role of feature extractor.

In addition to methodological differences, AEFE shows supplementary strengths compared to FM. Benefit from the setting of the time windows, the features constructed by AEFE can change over time on the grounds that it utilizes timing information to treat the historical samples in a discriminative manner. Moreover, AEFE does not only use \emph{label} as indicator and \emph{sum/mean} as operator. When more elements are added to each set, AEFE can mine more data patterns. Conversely, the analogy of Figure \ref{fig:aefevsfm} also helps to contemplate the use of representation learning to capture more forms of features similar to ones constructed in our framework without using \emph{label}.

\section{Optimization \& Implementation}
\label{sec:OptimizationImplementation}

Reasonable optimization techniques and efficient implementation are necessary for an automatic feature framework to be used under acceptable resource constraints. In this section, we will discuss the optimization and implementation of AEFE. First, optimization techniques include data sampling and Matrix Factorization-based field combination search are described in Section \ref{subsec:datasampling} and Section \ref{subsec:mf-basedFCS}. Then, a distributed implementation will be introduced in Section \ref{subsec:distributedimplementation}.

\subsection{Data Sampling}
\label{subsec:datasampling}

Prior to feature construction, proportional sampling will be performed on the raw dataset. These sampled data are used for successive phases to generate the valid combinatorial features list. Finally, the complete dataset is reused to generate the new dataset.

The primary purpose of data sampling is to make our proposed framework suitable for large datasets since the groupby-then-aggregating operation in the feature construction phase brings a large memory overhead. At the same time, data sampling can make related phases faster. However, a low sampling rate may lead to the problem of constructing features ineffective, so choosing the appropriate sampling rate is a trade-off between efficiency and effectiveness.

There are some works to apply sampling on similar problems, such as the automatic feature frameworks OneBM \cite{lam2017oneBM} and Cognito \cite{khurana2016cognito}, which can also be used on large datasets, but these works do not carry out experimental analysis on the impact of sampling. Additionally,  some machine learning models use a negative down-sampling strategy on CTR tasks to increase speed and slightly improve accuracy \cite{he2014practical_GBDT2LR}. Although we may also face the class imbalance problem, we do not conduct negative down-sampling, because it would seriously damage the original distribution and affect the calculation of indicators such as \emph{mean}.

\subsection{MF-based Field Combination Search}
\label{subsec:mf-basedFCS}

Assuming that the time complexity of constructing a feature is $C$, which is related to the sample size, then, from Equation \ref{eq:feature_num} and Section  \ref{subsec:featureconstruction}, the time complexity of the whole feature construction phase is $O(Q\times C)$ or $O(|\mathcal{F}|^2\times |\mathcal{I}|\times |\mathcal{W}|\times |\mathcal{O}|\times |\mathcal{P}|\times C)$. Most of the variables are in single digit, except for the number of categorical fields $|\mathcal{F}|$, which is generally tens(then $|\mathcal{F}|^2$ is hundreds). In order to reduce the time consumption, the search for categorical fields is most needed for optimization. Hence we use a heuristic strategy, named Matrix Factorization-based field combination search, to search for valid feature combinations as much as possible, avoiding wasting time on searching for invalid feature combinations.

The core idea of our proposed search strategy is to model the combining ability of a categorical field (hereafter called "field"), which is defined as the effectiveness of combinatorial features generated from this field with other fields. The best-performing field pair is selected in each step for feature constructing. After that, the parameters are updated by evaluating the validity of features. To do this, the following assumptions are made:
\begin{enumerate}
    \item The combining ability of a field is migratable. That is, if the combinatorial features of field A and B work well, the combination of field A and other fields is more likely to be effective;
    \item The combining ability of features generated by different operations is independent. That is, if field A and B get a positive result with \emph{mean} as the operator, it does not mean that operator \emph{std} is equally good for field A and B.
\end{enumerate}

\begin{algorithm}[t]
    \SetAlgoLined
    \SetKwInOut{Input}{Input}\SetKwInOut{Output}{Output}
    \SetKwData{maxIterations}{maxIterations}
    \SetKwFunction{MF}{MF}\SetKwFunction{FSA}{FSA}

    \Input{$\mathcal{D}$, $\mathcal{F}$}
    \Output{$\mathcal{G}$}
    initialize $\Bar{M}_{FP}\in \mathbb{R}^{m\times m}$ using Information Gain ratio(as Equation \ref{eq:MFP}, Equation \ref{eq:pij} and Equation \ref{eq:minmax})\;
    $V \leftarrow \MF(\Bar{M}_{FP}$)\;
    $\mathcal{G}\leftarrow \emptyset$\;
    $\maxIterations \leftarrow |\mathcal{F}|*(|\mathcal{F}|-1)/2$\;
    \For{$iter = 1$ to $\maxIterations$}{
        $i,j \leftarrow \operatorname*{arg\,max}_{i',j'} (p_{i'j'}) $\;
        Feature constructing using $F_i$ and $F_j$, and generated feature set $\mathcal{G}_{FG}$ obtained\;
        $N_{FG} \leftarrow |\mathcal{G}_{FG}|$\;
        % Feature selection phase and effectiveness feature set $\mathcal{G}_{valid}$ obtained\;
        $\mathcal{G}_{valid} \leftarrow \FSA(D, \mathcal{G}_{FG})$; \tcp*[f]{Feature selection phase}\label{FSA} \;
        $N_{valid} \leftarrow |\mathcal{G}_{valid}|$\;
        $\dot{p_{ij}}$ $\leftarrow N_{valid}/N_{FG}$\;
        Update $V$ by Equation \ref{eq:mfsgd1} and Equation \ref{eq:mfsgd2}\;
        $\mathcal{G}\leftarrow \mathcal{G}\cup \mathcal{G}_{valid}$\;
        $score_{new} \leftarrow$ evaluation score of $\mathcal{G}$\;
        \If{$score_{new}>score_{max}$}{
        $score_{max} \leftarrow score_{new}$\; 
        }
        \If{$score_{max}$ does not improve within $iter_{es}$ iterations}{
        break;
        }
    }
    \Return $\mathcal{G}$
    \caption{AEFE with MF-based Field Combination Search\label{alg:AEFE_MF}}
\end{algorithm}

The combining ability of a field is quantifiable under Assumption 1, while Assumption 2 is made for the distributed implementation as mentioned in Section  \ref{subsec:distributedimplementation}. Our proposed AEFE with MF-based field combination search can be briefly described in Algorithm \ref{alg:AEFE_MF} and the details are as follows:
\begin{enumerate}
    \item Latent vectors $\mathbf{v}_{i}$ represents the combining ability of $F_i$, and the effectiveness of combinatorial features of $F_i$ and $F_j$ is expressed as $p_{ij}=\mathbf{v}_{i}\cdot \mathbf{v}_{j}$, then the matrix of all field combinations is $\Bar{M}_{FP}$:
    \begin{equation}
        \Bar{M}_{FP} = 
             \begin{pmatrix}
              p_{11} & p_{12} & \cdots & p_{1m} \\
              p_{21} & p_{22} & \cdots & p_{2m} \\
              \vdots  & \vdots  & \ddots & \vdots  \\
              p_{m1} & p_{m2} & \cdots & p_{mm} 
             \end{pmatrix} \label{eq:MFP}
    \end{equation}
    $\Bar{M}_{FP}$ is a symmetric matrix with diagonal elements $p_{ii}=0$. In order to find the initial value of $\mathbf{v}$, $p_{ij}$ is initialized by the \emph{Information Gain Ratio}, specifically:
    \begin{equation}
        \Bar{p_{ij}}=\Bar{p_{ji}}=\frac{\sum_{l\in L}\sum_{c\in C_{ij}}p(c,l)log[\frac{p(c,l)}{p(c)p(l)}]}{H(C_{ij})}, \quad i\neq j \label{eq:pij}
    \end{equation}
    \begin{equation}
        p_{ij}=MinMaxScaler(\Bar{p_{ij}}) \label{eq:minmax}
    \end{equation}
    Here $C_{ij}$ is the Cartesian product of the features in fields $i$ and $j$, $L$ is the set of target values (\emph{labels}), and $H(X)$ represents \emph{entropy} of the random variable $X$. After calculating the information gain ratio of all elements, the entire matrix is transformed into $[0,1]$ interval.
    
    \item Matrix factorization of $\Bar{M}_{FP}$ yields the matrix $V=MF(\Bar{M}_{FP})=[\mathbf{v}_1,\cdots,\mathbf{v}_m]$, and $V$ roughly satisfies the equation $\Bar{M}_{FP}=V^TV$. Then the latent vectors representing the combining ability of $F$  are obtained.
    
    \item Select the largest $p_{ij}$ that is unused for combination in matrix $M_{FP}$, and use $F_i$ and $F_j$ to construct features with other elements(operators, indicators etc.). The total number of combinatorial features for these two fields is $N_{FG}$.
    
    \item The number of combinatorial features that can be retained after the feature selection algorithm is $N_{valid}$; then, we regard the \emph{real} effectiveness of combinatorial features of $F_i$ and $F_j$ is $\dot{p_{ij}}=N_{valid}/N_{FS}$.
    
    \item Following the idea of iterative computation matrix factorization \cite{lee2001NMF}, with the current fields pair as a sample, $p_{ij} , \dot{p_{ij}}$ are the predicted value and the target value, respectively, and $ \mathbf{v}_{i}$ and $\mathbf{v}_{j}$ are updated by gradient descent:
    \begin{align}
        v_{ik}^{t+1}& \leftarrow v_{ik}^{t} - \eta(loss(\dot{p_{ij}}-p_{ij}))v_{jk}^{t} \label{eq:mfsgd1} \\
        v_{jk}^{t+1}& \leftarrow v_{jk}^{t} - \eta(loss(\dot{p_{ij}}-p_{ij}))v_{ik}^{t} \label{eq:mfsgd2} 
    \end{align}
    Where $v_{ik}^{t}$ represents the $k$th component of $\mathbf{v}_{i}$ in the $t$th iteration, and $\eta$ is the learning rate. By updating the latent vectors, the combining ability of $F_i$ and $F_j$ can be changed in time, so that field combination is directed to the highest yield in each iteration.
    \item Repeat Step 3 to Step 5 until the termination condition, which is the evaluation metric not improving within continuous $iter_{es}$ iterations or completing all field combinations.
\end{enumerate}

\subsection{Distributed Implementation}
\label{subsec:distributedimplementation}

Multiple feature selection and MF-based field combination search make AEFE feasible. Further, we propose a distributed implementation scheme using task-parallelism to improve the speed of the framework.

The most time-consuming phase of the whole framework is to obtain the valid combinatorial features list through feature combination and feature selection, so we divide it into a combination of operator set and indicator set, and acquire $N_{task}<|\mathcal{I}|\times |\mathcal{O}|$ tasks(Partial aggregation functions are not meaningful for some indicators, so the total number of tasks will be less than $|\mathcal{I}|\times |\mathcal{O}|$). These tasks are assigned to different compute nodes.

Each node independently performs the two main phases - \emph{Feature Construction, Feature Selection} - of AEFE. The hypothesis that the combining ability of features generated by different operations is independent supports MF-based field combination search can be used independently after task partitioning. In addition, due to the larger number and the greater difference of the candidate features constructed by each task are summarized and there is still redundancy, \emph{Global Feature Selection} need to perform, which is similar to Algorithm \ref{alg:FSA}.

\section{Experiment}
\label{sec:Experiment}

In this section, we will first introduce the experiment settings in Section  \ref{subsec:ExperimentSetting}, and then answer the following questions through a series of experiments:
\begin{itemize}
    \item \textbf{(RQ1)} How much improvement does our proposed AEFE bring to the models, and how does our method perform compared to state-of-the-art deep learning models?
    \item \textbf{(RQ2)} Can AEFE effectively capture field combination information? What is the interpretability of this captured information?
    \item \textbf{(RQ3)} What are the effects of two optimization techniques---data sampling and MF-based field combination search?
\end{itemize}

\subsection{Experiment Setting}
\label{subsec:ExperimentSetting}

\subsubsection{Datasets}
\label{sssec:Datasets}

We conduct experiments on two public datasets and one private dataset, which are ad click-through rate prediction datasets. The summary statistics of three datasets are shown in Table \ref{table:dataset}.

\begin{table}[ht]
    \caption{Statistics of datasets. \label{table:dataset}}
    \centering
    \begin{tabular}{llll}
    \toprule
    Dataset & \#samples & \#Fields & \#Features \\ \midrule
    Iqiyi   & 22,520,404      & 16       & 4,926,259  \\ 
    Ali     & 26,557,961      & 21       & 2,960,015  \\ 
    Avazu   & 40,428,967      & 22       & 1,149,920  \\ 
    \bottomrule
    \end{tabular}
\end{table}

\begin{itemize}
    \item \textbf{Iqiyi}: A private online video display ad dataset. A 20-day impression log of an advertising platform on Iqiyi online video website from July 4, 2016, to July 23, 2016. The final dataset is obtained by pre-processing the deletion of missing values, fraudulent click record filtering, and the like. 
    \item \textbf{Ali}\footnote{\url{https://tianchi.aliyun.com/dataset/dataDetail?dataId=56}}: A public e-commerce platform displays advertising data sets. We use \texttt{raw\_sample} as the main table, and \texttt{raw\_behavior\_log} is simply processed and connected to the main table to get \textit{btag, cat, brand} 3 fields, then right join \texttt{ad\_feature, user\_profile} with the main table to get the advertising and user profile information for each log.
    \item \textbf{Avazu}\footnote{\url{ https://www.kaggle.com/c/avazu-ctr-prediction}}: A Kaggle competition dataset contains the click log from October 21st to 29th, 2014 (because the logs on October 30th have no label information, so they are dropped), we group the features that appear less than 10 times in a field to one feature, and then perform some other simple preprocessing.
\end{itemize}
We uniformly use the last day's log as a test set and all the previous data as a complete training set. In order to select the best model hyper-parameters,  $1/5$ data were randomly divided from the training set as the validation set for hyper-parameter tuning.

\subsubsection{Evaluation Metrics}
\label{sssec:EvaluationMetrics}

\textbf{AUC} (Area Under ROC) is adopted as the main metric for evaluating the performance of the models. AUC can be understood here as the probability of correct ranking of a random “positive”-“negative” pair. AUC is a commonly used classification task evaluation indicator because it is not sensitive to class imbalance and can better reflect the needs of the practical scene. If the class of a sample is randomly predicted, the value of AUC is 0.5. In addition, for the purpose of clearly comparing the performance differences between our method and other comparison models, the Relative Improvement of AUC (\textit{RelaImpr}) \cite{yan2014CGL} will also be shown. The expression of \textit{RelaImpr} is as follow:
\begin{equation}
    relaImpr=[\frac{AUC(Model)-0.5}{AUC(Baseline)-0.5}-1]\times 100\% \label{eq:relaimpr}
\end{equation}

\subsubsection{Baseline Models}
\label{sssec:Baseline Model}

To observe the performance of AEFE, we chose some classical machine learning models and state-of-the-art deep learning models in the same problem domain as the baseline models, and most of them modeling feature crossing. Other automatic feature engineering frameworks are not compared because they are not suitable for a dataset that is almost all categorical fields. The baseline models are introduced as follows:
\begin{itemize}
    \item \textbf{Logistic Regression(LR)}:  A simple linear model that can only learn the weight of each independent feature. It is widely used to handle sparse categorical features because of its simplicity. We add L2 regularization to avoid overfitting. 
    \item \textbf{LightGBM(GBDT)} \cite{ke2017lightgbm}: GBDT can learn feature crossing of any order. LightGBM is an efficient implementation of GBDT, and it supports categorical features well. We implements it using \cite{ke2017lightgbm}'s code\footnote{\url{https://github.com/Microsoft/LightGBM}}.
    \item \textbf{Factorization Machine(FM)} \cite{rendle2010factorization}: A linear model capable of learning second-order feature crossing. It is generally applied for recommender system, etc. It is the basis of many deep learning models in these domains since it can learn the embedding vectors with feature crossing information.
    \item \textbf{Field-aware Factorization Machine(FFM)} \cite{juan2016fieldFFM}: It extends the embedding vector representation of FM, and features interact with different fields' features using different embedding vectors. We used LibFFM package\footnote{\url{https://github.com/guestwalk/libffm}} in our experiments, and for the Avazu dataset, we use the hyper-parameters of the original paper: $learningrate=0.2,L_{FFM}=0.00005,k_{emb}=4$.
    \item \textbf{Attentional Factorization Machines(AFM)} \cite{xiao2017AFM}: A variant of the FM model that introduces the attention mechanism. It can learn different weights for different crossing features.
    \item \textbf{Neural Factorization Machines(NFM)} \cite{he2017NFM}: NFM generalizes FM to deep learning model, using the Bi-interaction layer and fully-connected layers to learn higher-order nonlinear interactions between features.
    \item \textbf{Deep Factorization Machines(DeepFM)} \cite{Guo2017DeepFM}: An improved version of Wide\&Deep model that uses FM and DNN to train the embedding layers in parallel and can take advantage of both second-order and higher-order feature crossing information.
    \item \textbf{eXtreme Deep Factorization Machines(xDeepFM)} \cite{lian2018xdeepfm}: A network structure that can learn implicit and explicit feature crossing by using DNN and Compressed Interaction Network (CIN) at the same time. We choose this model as a representative of non-FM-embeddings deep learning models for comparison.
\end{itemize}

\begin{table}[ht!]
    \caption{Grid search scope of model hyper-parameters.}
    \label{table:gridsearch}
    \centering
    \begin{tabular}{ll}
        \toprule
        Model       & hyper-parameter   \\ \midrule
        LR          & $L_{LR}=[10^{-3},10^{-4},10^{-5},10^{-6}]$    \\
        FM          & \makecell[tl]{$k_{emb}=[10,20,30]$ \\ $Dt_{FM}=[0.3,0.6,0.9]$ }    \\
        FFM         & \makecell[tl]{$k_{emb}=[4,6,8]$ \\ $L_{FFM}=[10^{-3},10^{-4},10^{-5}]$ \\ $learningrate=0.2$}  \\
        AFM         & \makecell[tl]{$k_{emb}$ follow the best result of \emph{FM} \\ $k_{att}=256$ \\ $L_{AFM}=[0.1,1,10]$ \\ $Dt_{AFM}=[0.3,0.6,0.9]$}   \\
        NFM         & \makecell[tl]{$k_{emb}$ follow the best result of \emph{FM} \\ $net_{DNN}=[(200),(200\times2),(200\times3)]$ \\ $Dt_{DNN}=[0.3,0.6,0.9]$ \\ $L_{DNN}=[10^{-3},10^{-4},10^{-5}]$}  \\
        DeepFM      & \makecell[tl]{$k_{emb}$ follow the best result of \emph{FM} \\ $net_{DNN}=[(200),(200\times2),(200\times3)]$ \\ $Dt_{DNN}=[0.3,0.6,0.9]$ \\ $L_{DNN}=[10^{-3},10^{-4},10^{-5}]$}  \\
        xDeepFM     & \makecell[tl]{$k_{emb}$ follow the best result of \emph{FM} \\ $net_{DNN}=[(200),(200\times2),(200\times3)]$ \\ $net_{CIN}=[(200),(200\times2),(200\times3)]$ } \\
        GBDT         & \makecell[tl]{$reg_{\alpha}=[0.01,0.1,1]$ \\ $reg_{\beta}=[1,10,100]$ \\ $learningrate=0.05$ }    \\
        \bottomrule
    \end{tabular}
    \begin{center}
        Annotate: $L_{*}$=L2 regulization coefficient on specific parameters, $k_{emb}$=embedding size,$k_{att}$=attention factors of AFM, $Dt_{*}$=dropout ratio of specific layers, $net_{*}$=strucure of specifict network, $reg_{\alpha},reg_{\beta}$=L1,L2 regulization coefficient of GBDT
    \end{center}
\end{table}

We use AEFE-generated features to train LR and GBDT (denoted as \textbf{AEFE+LR} and \textbf{AEFE+GBDT}, respectively) against the above model to verify the validity of AEFE. For fairness, the key hyper-parameters of these models are determined by grid search, and the hyper-parameters with the best effect on the validation set are selected for testing. The hyper-parameter search scope is shown in Table  \ref{table:gridsearch}. Except for the model that explicitly specifies the implementation, other models are implemented using Tensorflow\footnote{\url{https://www.tensorflow.org}}, and the optimization method is mini-batch Adam\cite{kingma2015adam} with learning rate selected from $0.0001$ to $0.0005$.

\subsubsection{Experiment Parameters \& Environment}
\label{sssec:Environment}

\begin{table}[ht!]
\caption{Main parameters of AEFE.}
\label{table:framework_parameters_setting}
\centering
\begin{tabular}{@{}lccc@{}}
    \toprule
    Parameters         & Iqiyi           & Ali         & Avazu        \\\midrule
    Sampling rate      & $0.1$             & 0.05           & 0.02      \\
    Time windows(days) & $(3,5,7,14)$        & $(3,5)$         & $(3,5)$ \\
    Operator set       & \multicolumn{3}{l}{$sum,mean,count,max,min$} \\
    Paradigms          & \multicolumn{3}{l}{4 paradigms in Table \ref{table:paradigmandfeature}}                       \\ \bottomrule
\end{tabular}
\end{table}

The main parameters of AEFE in the experiments are shown in Table  \ref{table:framework_parameters_setting}. Time windows is determined by the time span of the dataset while operator set and construction paradigms are fixed to the same for all dataset. Sampling rate will be discussed in Section \ref{sssec:samplerateanalysis}. In addition, XGBoost \cite{chen2016xgboost} is implemented in Algorithm \ref{alg:FSA} (FSA) as the learning model. Meanwhile $t_{filter}$, $t_{embbed}$, and $t_{wrapper}$  are set as $0.00001$, $0.02$, and $0$ respectively.

AEFE is implemented and experimented with Python 3.6. It is deployed on an 8-node platform with each node configured as CPU: Intel Xeon E5-2670 @2.6Ghz*16core*2, memory: 128GB. Other models implemented by non-Tensorflow are also running on these nodes.

The version of Tensorflow for the deep learning model is r1.12.0. The machine used is configured as CPU: Intel Xeon e5-2603 v4.@1.70ghz, GPU: NVIDIA GTX 1080Ti*2, and memory: 64GB.

\subsection{Performance Comparison}
\label{subsec:PerformanceComparison}

\subsubsection{Comparison with Baseline Models(RQ1)}
\label{sssec:ComparisonWithBaselineModels}

The performance of different models on the test set is summarized in Table \ref{table:performance_results}, which contains the evaluation metric AUC and \textit{ReleImpr}(original features vs. AEFE's features). All results are the average of 10 replicates.

\begin{table}[ht]
    \caption{Performance(AUC) comparison results.}
    \label{table:performance_results}
    \centering
    \begin{tabular}{@{}llll@{}}
    \toprule
    Model     & Iqiyi & Ali           & Avazu             \\ \midrule
    LR        & 0.58087(\textit{41.29\%}) & 0.64780(\textit{67.83\%}) & 0.73122(\textit{10.25\%})\\
    GBDT      & 0.56330(\textit{80.51\%}) & 0.69172(\textit{29.39\%}) & 0.74908(\textit{2.34\%})\\
    FM        & 0.61221(\textit{1.83\%}) & 0.67355(\textit{42.93\%}) & 0.73998(\textit{6.22\%}) \\
    FFM       & 0.60473(\textit{9.10\%}) & \underline{0.71449(\textit{15.65\%})} & 0.75049(\textit{1.76\%}) \\
    AFM       & 0.60605(\textit{7.74\%}) & 0.65486(\textit{60.18\%}) & 0.74213(\textit{5.28\%}) \\
    DeepFM    & \underline{0.61254(\textit{1.53\%})} & 0.71159(\textit{17.24\%}) & 0.73539(\textit{8.29\%}) \\
    NFM       & 0.61252(\textit{1.55\%}) & 0.67441(\textit{42.23\%}) & 0.74432(\textit{4.33\%}) \\
    xDeepFM   & 0.60912(\textit{4.71\%}) & 0.71256(\textit{16.70\%}) & \underline{0.75258(\textit{0.92\%})} \\
    AEFE+LR   & 0.60540(\textit{8.41\%}) & 0.72232(\textit{11.58\%}) & 0.74571(\textit{3.74\%})          \\
    AEFE+GBDT & \textbf{0.61426} &   \textbf{0.74806} & \textbf{0.75491} \\
    \bottomrule
    \end{tabular}
    \begin{center}
    Note: bold indicates the best result, and underline indicates the best result of baseline models. The  content in italics in parentheses indicates the improvement of AEFE+GBDT relative to this model on \textit{RelaImpr}.
    \end{center}
\end{table}

\begin{table}[ht]
    \caption{P-value of t-test on AEFE+GBDT vs. other models.}
    \label{table:pvalue}
    \centering
    \begin{tabular}{@{}llll@{}}
    \toprule
    Comparison Model                            & Iqiyi         & Ali           & Avazu             \\ \midrule
    \makecell[tl]{LR, GBDT, FM,\\FFM, AFM, NFM} & $<10^{-6}$    & $<10^{-6}$    & $<10^{-6}$ \\
    DeepFM                                      & $<10^{-3}$    & $<10^{-6}$    & $<10^{-6}$ \\
    xDeepFM                                     & $<10^{-6}$    & $<10^{-6}$    & $<10^{-3}$ \\    
    \bottomrule
    \end{tabular}
\end{table}

The following points are known from Table \ref{table:performance_results}:
\begin{itemize}
    \item Features generated by AEFE make LR and GBDT superior to those trained with raw features, which shows that our proposed framework can significantly enhance the performance of machine learning models. Especially on the Iqiyi dataset, the effect of LR or GBDT alone is not satisfying, while AUC can be sharply improved by using AEFE's combinatorial features.
    \item  AEFE+GBDT outperforms state-of-the-art models in three datasets. Compared to the best baseline, \emph{RelaImpr}s are $1.53\%$, $15.65\%$ and $0.92\%$ respectively. Meanwhile, we conduct 10 repeated experiments on baseline models and AEFE+GBDT for t-test, and the p-value is shown in Table \ref{table:pvalue}. The results show that the increase of AEFE on each dataset is significant.
    \item FM, which can capture the second-order feature interaction information, brings about obvious improvement compared with LR. However, although various improved models of FM have a larger \emph{Hypothesis Space} (sufficient to cover the expression of FM), they do not achieve the effect of FM on some datasets in our experiments. For example, AFM is worse than FM on Iqiyi and Ali datasets. On the contrary, our method is more stable in all experiments.
    \item AEFE+LR, using the simplest classifier, outperforms all baseline models on the Ali dataset, which shows that effective constructed features are more powerful than a complex learning model in some datasets of such tasks.  This is why we still need to focus on automatic feature engineering.
\end{itemize}

Experiments in this section show that our proposed AEFE improves the performance of models significantly and AEFE+GBDT achieves better AUC compared with the state-of-the-art deep learning models.

\subsubsection{Comparison on Field Combination(RQ2)}
\label{sssec:Comparison_on_fields_combination}

In addition to the accuracy comparison with deep learning models,  we compare the capabilities to capture second-order feature interaction of AEFE and FM in this section. Then we study the interpretability of the features generated by AEFE.

To analyze the effectiveness of the features interaction of different field pairs, we define the \emph{Combination Strength Matrix}($CS$) of AEFE and FM respectively: for our proposed AEFE+GBDT, GBDT can easily output the feature weights, and then we divide and accumulate the feature weights by field combination to get the combined strength of different fields for the prediction, as in the form of Equation \ref{eq:cs_aefe}; for FM, which cannot directly obtain the importance of field combination, we take the practice of calculating the field \textit{mean embedding} in \cite{qu2018pin}'s paper. The vector of field $i$ is represented as $\Bar{v}_i=\sum^{d_i}_{j=1}v^j_i/d_i$, the field combination strength expression is shown in Equation \ref{eq:cs_fm}. Different from that paper, since we need to know the importance of the field combination to the prediction without distinguishing its positive and negative, we will take the absolute value of the matrix.

\begin{align}
    CS^{AEFE}_{(i,j)}&=\sum\limits_{g_k \, belong \, (i,j)} w_k \label{eq:cs_aefe} \\
    CS^{FM}_{(i,j)}&=|\Bar{v}_i\cdot\Bar{v}_j| \label{eq:cs_fm}
\end{align}

\begin{figure}
    \centering
    \includegraphics[width=\linewidth]{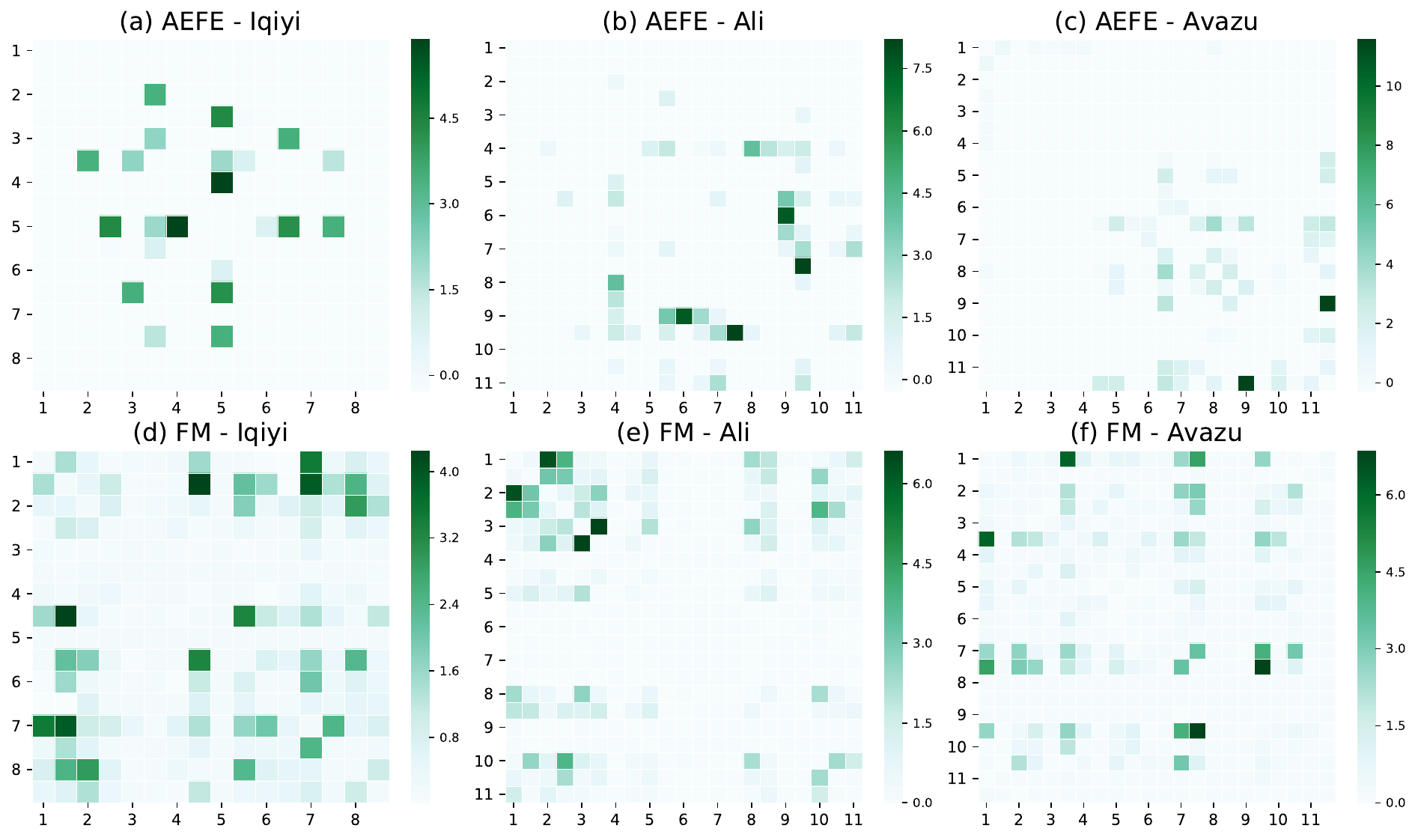}
    \caption{Field combination heat map of AEFE and FM on Iqiyi, Ali and Avazu datasets. Darker represents stronger.}
    \label{fig:aefevsfm_heatmap}
\end{figure}

As \cite{qu2018pin} mention, we can assume that the interaction of fields should be sparse. So, ideally, only a few values are larger and mostly smaller in the combination strength matrix. We plot the combination strength matrix of AEFE and FM into heat maps(Figure \ref{fig:aefevsfm_heatmap}). The two rows from top to bottom are the results of AEFE and FM, with field number on the x-axis and y-axis. The grid $(i,j)$ represents the strength of the field combination of $F_{i}$ and $F_{j}$. The darker the color, the higher the intensity.

The experimental results show that only a few field combinations can have higher intensity, and both methods are consistent with the "hypothesis" of sparseness. On the one hand, it shows that the combinatorial features are of considerable importance to the model. On the other hand, the number of field combinations that contribute to the prediction is limited. Besides, AEFE's strength matrix is more sparse than FM's, because many invalid field combinations are directly discarded by AEFE, and their weight is $0$. Finally, we find an impressive phenomenon from the comparison of the heat maps: the important field combinations they capture rarely overlap. Therefore, we use Ali dataset, a public dataset with a clear meaning for each field, as an example for further analysis.

\begin{table}[ht]
    \centering
    \caption{Fields of the Ali dataset. \label{table:field_names}}
    \begin{tabular}{@{}lll@{}}
    \toprule
    Field Group & Field No. & Field Name                                                           \\ \midrule
    User profile & 1-9       & \makecell[tl]{final\_gender\_code,  age\_level,  pvalue\_level, \\shopping\_level, occupation, new\_user\_class\_level, \\user, cms\_segid, cms\_group\_id}  \\
    Ad information & 10-15      & \makecell[tl]{adgroup\_id, ad\_cate, campaign\_id, customer, \\ad\_brand, price}    \\
    User behavior & 16-18     & btag, cate, brand  \\
    Others & 19-21     & pid, hour, minute \\ \bottomrule
    \end{tabular}
\end{table}

\begin{table}[ht]
    \centering
    \caption{Some fields' meaning of the Ali dataset. \label{table:field_Description}}
    \begin{tabular}{@{}ll@{}}
    \toprule
    Field Name              & Description                                                \\ \midrule
    ad\_brand               & brand ID of advertisement                                   \\
    ad\_cate                & category ID of advertisement                                \\
    brand                   & brand ID of item that user have interacted with recently    \\
    cate                    & category ID of item that user have interacted with recently \\
    final\_gender\_code     & user gender                                                 \\
    pvalue\_level           & user consumption level                                      \\
    new\_user\_class\_level & city level of user                                          \\
    occupation              & whether the user is a college student                       \\ \bottomrule
    \end{tabular}
\end{table}

Field names of the Ali dataset in Figure \ref{fig:aefevsfm_heatmap} are shown in  Figure \ref{table:field_names}, while the meaning and group of some fields are listed in Figure \ref{table:field_Description}. In Figure \ref{fig:aefevsfm_heatmap}, the field combination that contributes the most to the prediction in AEFE is \emph{cate $\times$ ad\_cate} and \emph{brand $\times$ ad\_brand}, i.e., the field pairs mined by AEFE are the combination of user behavior features and advertisement information features, which means that user's recent interaction record with a specific brand or a specific category of goods implies the user's interest in the current advertisement. However, FM considers that \emph{final\_gender\_code $\times$ pvalue\_level} and \emph{new\_user\_class\_level $\times$ occupation}---the intra-combination of user profile features---are most important, which means that more detailed user group division (such as gender and consumption grades should be jointly divided) can bring better personalized prediction. Although we cannot absolutely point out which combinations of fields in this scenario are better than another, generally, the combination of users and items is beneficial and needs to be fully mined in tasks like recommender systems. On the other hand, some industrial models with real-time requirements will deliberately omit intra-group combinations to reduce computational complexity because they tend to have less impact on accuracy. Therefore, we believe that AEFE can be more accurate in identifying potential data patterns at the field-level. As evidence of our judgment, our approach also results in $42.92\%$ \emph{RelaImpr} to FM.

\begin{table}[ht]
    \centering
    \caption{Weights of some features belonging to two important field pairs of the Ali dataset. \label{table:feature_importances}}
    \begin{tabular}{@{}ll@{}}
    \toprule
    Combinatorial Feature & weight   \\ \midrule
    $GA_{std}^{5ds}(ad\_cate\&cate,click)-GA_{std}^{5ds}(ad\_cate,click)$  & $2.35\%$ \\
    $GA_{std}^{5ds}(ad\_cate\&cate,click)$  & $2.26\%$ \\
    ${GA_{sum}^{5ds}(ad\_brand\&brand,click)}/{GA_{sum}^{5ds}(ad\_brand,click)}$  & $1.45\%$ \\
    ${GA_{sum}^{5ds}(ad\_brand\&brand,impression)}/{GA_{sum}^{5ds}(ad\_brand,impression)}$   & $1.38\%$ \\ \bottomrule
    \end{tabular}
\end{table}

Furthermore, with the visualization, the interpretability of learning models such as FM is limited to which field combinations are more important. AEFE can further reveal the paradigm of important combinatorial features. As shown in Table \ref{table:feature_importances}, there are four most weighted features of the above two fields in the Ali dataset. First, the first two important features are about the click sequence variance rather than the most intuitive click rate (\emph{click:mean}), which means that the divergence of click sequence of a user who recently interacted with a particular merchandise category and a particular ad category is a strong predictor. Second, 3 of the 4 features are complex ratios or difference features, indicating that this complicated paradigm is more effective. Finally, in this experiment, time windows are set to $[3,5]$, and the higher weights in Table \ref{table:feature_importances} are all 5 days, which means that a more extended time window is more suitable in this dataset.

The elaborated exploration of generated features above is a demonstration of AEFE as an expert and intelligent system at the analysis level. For one thing, it is helpful for mining important features to guide data collection and to preprocess. For another, it enlightens data analysts to conceive stronger features based on them---in fact, AEFE is not a complete replacement for the work of data scientists (this is a goal that other similar work cannot achieve now) but makes the work more efficient and inspiring.

\subsection{Optimization Analysis(RQ3)}
\label{subsec:Optimizationanalysis}

AEFE uses a series of optimization to improve its efficiency. We conduct two experiments to analysis whether these techniques can bring acceptable results. The classifier used in this section is GBDT.

\subsubsection{Sampling Rate Analysis}
\label{sssec:samplerateanalysis}

It is mentioned in Section \ref{subsec:datasampling} that data sampling is used to speed up feature constructing and save system resources. Obviously, the lower the sampling rate brings faster speed, but at the same time, it damages the effect. Our experiment focuses on the relationship between the sample rate and feature quality. Here we define \emph{Deviation of Weights' Gap}(DoWG) to measure whether a sampling rate is effective enough, the expression is as follows:
\begin{equation}
    r^{s}_{ij}=|\frac{|w^{(s)}_i-w^{(s)}_j|-|w^{(1.0)}_i-w^{(1.0)}_j|}{w^{(1.0)}_i-w^{(1.0)}_j+\epsilon}| \label{eq:DoWG}
\end{equation}
Where $w^{(s)}_i\in[0,1]$ denotes the weight of feature $i$ in the model trained with sample rate $s$, and $\epsilon$ is $10^{- 30}$ to avoid the denominator to be 0. Apparently, the closer $r^{s}_{ij}$ is to $0$, the more similar the relative importance of features with sample rate$s$ is to unsampling, and the better the simulation of this sample rate is for the whole dataset. In particular, $r^{ 1.0}_{ij}=0$.

\begin{figure}
    \centering
    \includegraphics[width=\linewidth]{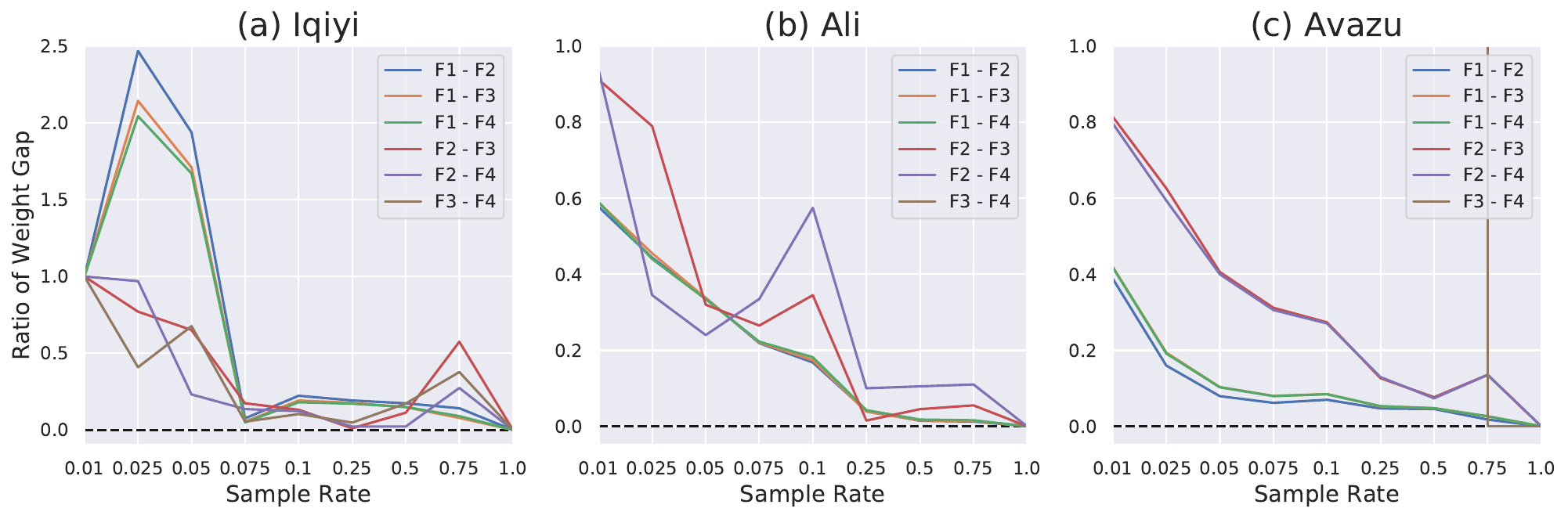}
    \caption{Comparison of different data sampling rate. The black dotted line represents the best fit. }
    \label{fig:samplerate}
\end{figure}

We sort the generated features in descending order of feature importance and select the first feature, the two trisection points, and the last one, labeled \textbf{F1, F2, F3, F4}. Then sample rates including $[0.01,0.025,0.05,0.075,0.1,0.25,0.5,0.75,1.0]$ are used for experiments. In order to make the experiments more stable, the results obtained at each sampling rate are the average of 10 repeated experiments. The experimental results are shown in Figure \ref{fig:samplerate}.

From Figure \ref{fig:samplerate}, we can observe the following phenomena: 1) The larger the sampling rate, the closer the DoWG is to 0, which is consistent with intuition. When the sampling rate reaches 0.25 or higher, the effect is close to perfect fitting. 2) Comparing the datasets with different sizes, it can be found that the larger the sample size, the lower the sampling rate requirement. Even with a sampling rate of $0.05$, the experiment result is fine on the Avazu dataset, but in the experiments of Iqiyi and Ali, there is still an apparent fluctuation at higher sample rates. 3)DoWG between features with lower weights (such as \textbf{F3-F4}) fluctuates wildly, which may result in the inability to pass feature selection due to randomness. However, the features with higher weights are stable and can be guaranteed to be retained.

A higher sample rate should be chosen to achieve good simulation results. However, we only want to ensure that the valid features can be maintained after feature selection, so the deviation of feature weights without severe distortion can meet our requirement. And the feature weights need to re-train after generation. Conversely, lower sampling rates are more efficient.

\subsubsection{MF-based Search Analysis}
\label{sssec:mfbasedsearchanalysis}

\begin{figure}[t]
    \centering
    \includegraphics[width=\linewidth]{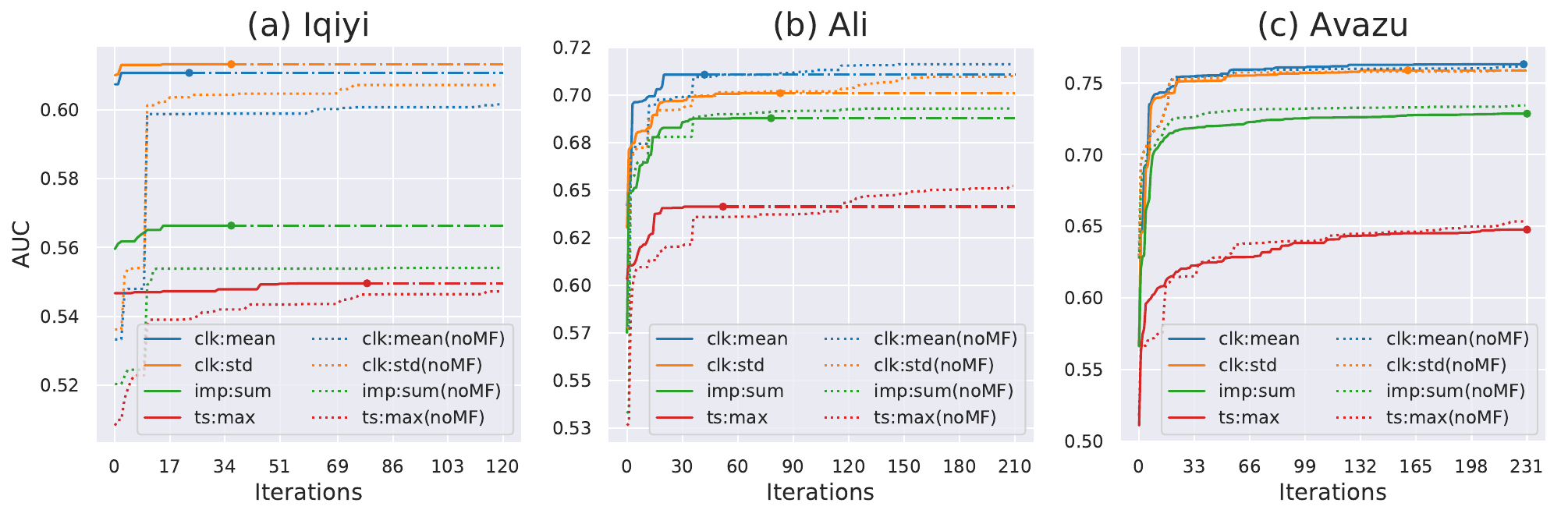}
    \caption{AUC changes for different tasks with or without MF-base search. Solid dots represent early stopping points. (clk, imp, and ts are short for click, impression and timestamp.)}
    \label{fig:mfvsaf}
\end{figure}

We design experiments to compare the performance of the complete AEFE and AEFE without MF-based field combination search to observe 1) whether this technique can improve efficiency, and 2) whether it affects accuracy. In Figure \ref{fig:mfvsaf}, there are AUC change curves of single-node tasks of generating effective combinatorial features lists. We have selected 4 out of 8 tasks, and their indicators and operators are \emph{click:mean}, \emph{click:std}, \emph{impression:sum} and \emph{timestamp:max} respectively. In addition, Table \ref{table:mfvsaf_statistic_iqiyi}, Table  \ref{table:mfvsaf_statistic_ali}, Table \ref{table:mfvsaf_statistic_avazu} contain comparisons of the results of the two methods.

\begin{table}[ht]
    \centering
    \caption{Statistic of AEFE with or without MF-based search on the Iqiyi dataset. \label{table:mfvsaf_statistic_iqiyi}}
    \begin{tabular}{@{}llll@{}}
    \toprule
    Method     & Avg. time & \#features & AUC     \\ \midrule
    AEFE(noMF) & 4.23h    & 25         & 0.61238 \\
    AEFE       & 2.37h     & 46         & 0.61416 \\ \bottomrule
    \end{tabular}
\end{table}

\begin{table}[ht]
    \centering
    \caption{Statistic of AEFE with or without MF-based search on the Ali dataset. \label{table:mfvsaf_statistic_ali}}
    \begin{tabular}{@{}llll@{}}
    \toprule
    Method     & Avg. time & \#features & AUC     \\ \midrule
    AEFE(noMF) & 4.20h    & 103        & 0.74546 \\
    AEFE       & 1.77h     & 119        & 0.74810  \\ \bottomrule
    \end{tabular}
\end{table}

\begin{table}[ht]
    \centering
    \caption{Statistic of AEFE with or without MF-based search on the Avazu dataset. \label{table:mfvsaf_statistic_avazu}}
    \begin{tabular}{@{}llll@{}}
    \toprule
    Method     & Avg. time & \#features & AUC     \\ \midrule
    AEFE(noMF) & 6.43h    & 232        & 0.75323 \\
    AEFE       & 4.75h    & 138        & 0.75607 \\ \bottomrule
    \end{tabular}
\end{table}

First, in Figure \ref{fig:mfvsaf}, we can see that most of AEFE's tasks terminate early, which reveals that MF-based search does improve efficiency. And from Table \ref{table:mfvsaf_statistic_iqiyi}, Table \ref{table:mfvsaf_statistic_ali}, Table \ref{table:mfvsaf_statistic_avazu}, column "Avg. time" demonstrates that by implementing MF-based search, the running time of each task is reduced, respectively, to $56.22\%$, $42.29\%$ and $73.96\%$. Then, the solid line representing AEFE almost always leads the dotted line representing AEFE(noMF) in the first ten rounds of iterations, indicating that MF-based search can find out stronger features faster. But in the experiments of Ali and Avazu, after AEFE(noMF) exhaustively searching from all combinatorial features, some tasks' AUC of AEFE(noMF) eventually exceed AEFE. Does it mean that AEFE(noMF) can perform better in the end? It can be seen from Table  \ref{table:mfvsaf_statistic_iqiyi}, Table \ref{table:mfvsaf_statistic_ali}, Table  \ref{table:mfvsaf_statistic_avazu} that AUC of AEFE is larger than AEFE(noMF), and more feature generated is not better. The reason may be that features that perform well on a single task probably have higher redundancy and are discarded in the global feature selection phase subsequently. What's more, it is worth noting that in some tasks, AEFE fails to trigger early stop on Avazu, and AEFE's single task AUC is not better than  AEFE(noMF)'s. However, AEFE achieves better AUC with less than 100 features compared to AEFE(noMF), which shows that even if the early stop mechanism is nearly ineffective, the good feature quality brought by MF-based search can make the model perform well.

\section{Conclusion and Future Works}
\label{sec:conclusion}

In this paper, we propose an automatic feature engineering framework for categorical features, named AEFE, which solves two fundamental problems: data sparsity and combinatorial feature construction, in e-commerce applications, and keeps good interpretability. In order to avoid time-consuming transversal overall combinatorial feature schemes, we propose an MF-based field combination search strategy. It effectively reduces about half of the running time and improves the accuracy of learning models. Furthermore, combined with data sampling, distributed implementation, and other technologies, AEFE becomes an efficient and easy-to-use feature framework.

Our method support decision-making at the application level and analysis level. It not only increases the prediction accuracy apparently even cascaded with shallow models like LR but also generates interpretable features for the follow-up data analysis. In fact, the custom paradigm, from a certain perspective, is a kind of domain knowledge and also a necessary factor for the interpretability of the complex generated feature while the proposed method is nearly data-driven. With small changes of custom configurations, AEFE can handle a large quantity of data as materials to generate features that are not easily constructed by humans and then select the most effective ones through specific criteria. In this way, our study is an attempt to combine domain-related knowledge with a data-driven method.

The experimental results show that AEFE cascaded with GBDT outperforms state-of-the-art deep learning models on several datasets. Compared to directly training LR or GBDT with raw features, they achieve a large relative improvement of AUC with features generated by AEFE. Through visual analysis, AEFE can mine different but valuable combinatorial features compared to FM. Further experiments confirmed the validity of data sampling and MF-based field combination search.

There are some limitations of our study for different reasons. Firstly, though higher-order interaction is helpful in some tasks, we only consider the combination of two categorical fields in AEFE because the complexity of the search space grows exponentially with the order. However, experiments reveal that this weakness can be compensated by GBDT, which learns feature interactions of any order, and outperform deep learning models that learn higher-order nonlinear interactions. Secondly, the scheme "Feature Engineering + Model Training" is not an end-to-end solution like deep learning models. But at present, we cannot tightly couple the entire process while maintaining equivalent interpretability. This could be an area of future research. Thirdly, the lack of experimental comparison of related automatic feature engineering work is another limitation. The main reason is that the related work investigated is not expert in handling such features. However, the state-of-the-art deep learning models used for comparison are influential and effective.

Aside from the potential research areas arising from the limitations of the current study, we would like to explore this work in three more directions next. The first is how to combine AEFE with FM and other FM-embedding deep learning models. As mentioned in Section \ref{sssec:featureextractorpers}, AEFE and FM capture utterly different information about field combinations. Therefore prediction accuracy may be improved by fusing the two methods properly. The general model ensemble may be a rough but effective way, but there are more suitable approaches left to explore. The second is to study how to design representation learning modules for learning more rich forms of constructed features (such as timestamp-type features generated by AEFE) for better model performance. Most of the deep learning models currently only capture the relation of original features and labels, which is not enough for the data mining tasks we are concerned with. The third is to use meta-learning in automatic feature engineering for categorical features so that the feature generation capabilities can be transferred to more unseen datasets. To archive this, we need to determine how to form meta-features and how to design the meta-model, which requires a further understanding of the deeper connotations of categorical features.

\bibliographystyle{unsrt}
\bibliography{references}

\end{document}